\documentclass{article}

\usepackage{microtype}
\usepackage{amssymb}
\usepackage{graphicx}
\usepackage{subcaption}
\usepackage{booktabs}
\usepackage{amsmath, amsthm} % used for boldsymbol.
\renewcommand{\vec}[1]{\boldsymbol{#1}} % Uncomment for BOLD vectors.
\usepackage{hyperref}

\usepackage[accepted]{icml2019}
\usepackage{amsfonts}       % blackboard math symbols
\usepackage{nicefrac}       % compact symbols for 1/2, etc.

\usepackage{algorithm, algorithmic}
\usepackage{tabularx}
\icmltitlerunning{Hyper-Sphere Quantization: Communication-Efficient SGD for Federated Learning}

\renewcommand{\vec}[1]{\boldsymbol{#1}} % Uncomment for BOLD vectors.

\newtheorem{SGD_C}{Theorem}
\newtheorem{SGD_NC}[SGD_C]{Theorem}
\newtheorem{SGD_GD_NC}[SGD_C]{Theorem}

\newtheorem{unbias}{Lemma}
\newtheorem{comp_var}[unbias]{Lemma}
\newtheorem{comp_err}[unbias]{Lemma}

\newcommand{\norm}[1]{\left\lVert #1 \right\rVert}
\newcommand{\R}{\mathbb{R}}
\newcommand{\M}[1]{\vec{#1}}
\newcommand{\E}[1]{\mathbb{E}\left[ #1 \right]}
\newcommand{\bE}[1]{\mathbb{E}\big[ #1 \big]}

\newtheorem{asmp_smooth}{Definition}
\newtheorem{alpha_comp}[asmp_smooth]{Definition}

\newtheorem{asmp_cv}{Assumption}
\newtheorem{asmp_ncv}[asmp_cv]{Assumption}
\newtheorem{asmp_vb}[asmp_cv]{Assumption}

\begin{document}

\twocolumn[
\icmltitle{Hyper-Sphere Quantization: Communication-Efficient SGD\\ for Federated Learning}



\icmlsetsymbol{equal}{*}

\begin{icmlauthorlist}
\icmlauthor{Xinyan Dai}{cuhk}
\icmlauthor{Xiao Yan}{cuhk}
\icmlauthor{Kaiwen Zhou}{cuhk}
\icmlauthor{Han Yang}{cuhk}
\icmlauthor{Kelvin K. W. Ng}{cuhk}
\icmlauthor{James Cheng}{cuhk,hw}
\icmlauthor{Yu Fan}{hw}
\end{icmlauthorlist}

\icmlaffiliation{cuhk}{The Chinese University of Hong Kong}
\icmlaffiliation{hw}{Huawei Technologies Co. Ltd}

\icmlcorrespondingauthor{Eee Pppp}{ep@eden.co.uk}

\icmlkeywords{Machine Learning, ICML}

\vskip 0.3in
]




\begin{abstract}
	The high cost of communicating gradients is a major bottleneck for federated learning, as the bandwidth of the participating user devices is limited. Existing gradient compression algorithms are mainly designed for data centers with high-speed network and achieve $O(\sqrt{d} \log d)$ per-iteration communication cost at best, where $d$ is the size of the model. We propose hyper-sphere quantization (HSQ), a general framework that can be configured to achieve a continuum of trade-offs between communication efficiency and gradient accuracy. In particular, at the high compression ratio end, HSQ provides a low per-iteration communication cost of $O(\log d)$, which is favorable for federated learning. We prove the convergence of HSQ theoretically and show by experiments that HSQ significantly reduces the communication cost of model training without hurting convergence accuracy.        
\end{abstract}

%The high cost of communicating gradients is a major bottleneck in distributed machine learning and the problem is more severe for federated learning, in which the bandwidth of the participating devices is limited. Existing gradient compression algorithms are mainly designed for data centers with high-speed network and achieve $O(\sqrt{d} \log d)$ per-iteration communication cost at best, where $d$ is the size of the model. We propose hyper-sphere quantization (HSQ), a general framework that covers a family of gradient compression algorithms and can be configured to achieve a continuum of trade-offs between communication efficiency and gradient accuracy. At the high compression ratio end, HSQ provides a low per-iteration communication cost at $O(\log d)$ to facilitate communication-efficient federated learning. We prove the convergence of HSQ and discuss its connections to existing gradient compression algorithms. Experimental results show model training with HSQ converges smoothly and the total amount of communication is significantly reduced.  
\section{Introduction}\label{sec:intro} 
 
Machine learning usually solves the optimization problem $x^{\star}=\arg\min_{x\in \mathbb{R}^d} {f(x)}$, where $f(x)$ is usually the average loss over samples, to obtain the model parameter, where $d$ is the size of the model. Currently, stochastic gradient descent (SGD)~\cite{sgd} is the most popular algorithm for this purpose, especially for training deep neural networks. Given an unbiased stochastic gradient $g$ such that $\mathbb{E}[g(x)]=\nabla f(x)$, SGD iteratively updates the model by     
\begin{equation}\label{equ:mips}
x_{t+1} = x_{t}-\eta_t g(x_{t}),
\end{equation}
where $\eta_t$ is the learning rate and $x_t$ is the model parameter at the $t$-th iteration. 

%The cost function $f(x)$ is usually the average cost on individual training samples $f(x) = \frac{1}{n}\sum_{i=1}^{n}f(a_i, x)$, in which $a_i$ is the $i$-th training sample. Therefore, the stochastic gradient can be calculated using a single (or a mini-batch) training sample as $g(\vec{x})=\nabla f(\vec{a}_i, \vec{x})$.

\textit{Federated learning} is an emerging machine learning paradigm in which many user devices (e.g., tablets, smart phones, also called clients) cooperate to train a model~\cite{fedlearning,federated,fedAverage}. In the typical setting of federated learning, user devices calculate gradients (or local updates, we use gradient to refer to stochastic gradient for conciseness) on their local samples and transmit the gradients to a central coordinator for model update. Federated learning is gaining increasing attention thanks to its unique advantages over data-center-based training~\cite{paramtersever,ring}: sensitive user data do not need to be uploaded to a data center, which better motivates users to participate~\cite{selectiveSGD}. Moreover, labels for some supervised tasks (e.g., next word prediction) can be inferred naturally from use interaction and used efficiently for local training~\cite{fedAverage}.   

%the training workload is offloaded to a massive number of user devices so that investments for data centers can be reduced.

%More critically, transferring a large amount of data not only degrades the performance of concurrent user applications, but also drains the device battery, both of which would strongly discourage users from participation.

As modern models are usually large (e.g., millions or even billions of parameters for deep neural networks), communicating the gradient is a major bottleneck for federated learning, as user devices commonly use wireless networks and have limited bandwidth.  For federated learning, a low~\textit{per-iteration communication cost}~\footnote{We define per-iteration communication cost as the number of bits needed to communicate a gradient $g \in \mathbb{R}^d$.} is important because users who cannot afford the per-iteration cost will not participate in the training at all. On the contrary, if the per-iteration cost is low, more users will be willing to participate in federated learning. This also leads to higher flexibility for federated learning to access different user devices in different iterations during training, thereby reducing the amount of communication conducted by an individual user.
 
\begin{table}[]
	\caption{Per-iteration communication cost (in bits), assuming that a floating point number has 32 bits.}
	\label{tab:cost}
	\begin{center}
		\begin{sc}
			\fontsize{8}{9}\selectfont
			\begin{tabular}{cccccc}
				\toprule
				Name & SGD & TernGrad & SignSGD & QSGD & HSQ \\ \midrule
				Cost & $32d$ & $d\log3$ & $d$ & $\sqrt{d}\log d $ & $\log d $  \\ \bottomrule
			\end{tabular}
		\end{sc}
	\end{center}
\vspace{-6mm}
\end{table}

A number of gradient compression algorithms have been proposed to reduce the cost of gradient communication~\cite{qsgd,signsgd,terngrad,gradiveq,adacomp,scalablebatch,sparsified,distributed_mean,non_iid_data,decentralized,double_squeeze,gradient_push,sublinear_convex}. However, these algorithms are designed for data-center-based training (e.g., clusters connected via high-speed network) and the compression is not sufficient for federated learning given its stringent requirement on per-iteration communication cost. Table~\ref{tab:cost} lists the per-iteration communication cost of some representative algorithms. Generally, there is a trade-off between communication efficiency and gradient accuracy, where a lower communication cost is achieved by transmitting less accurate gradients. QSGD~\cite{qsgd} achieves the state-of-the-art per-iteration communication cost at $O(\sqrt{d}\log d)$, with the variance bound of gradient blown up by $\sqrt{d}$
~\footnote{Let $\nabla f(x)$ be the actual gradient, $g(x)$ and $\tilde{g}(x)$ be an unbiased stochastic gradient and its compressed approximation, respectively. If the original variance bound is $B$ (i.e., $\mathbb{E}\left[\norm{g(x) - \nabla f(x)}^2 \right]\le B$) and the variance after quantization can be bounded by $\mathbb{E}\left[\norm{\tilde{g}(x) - \nabla f(x)}^2\right]\le \beta B$, then the blow-up is said to be $\beta$.}. Some heuristics can give a lower communication cost, but they do not come with convergence guarantees. To facilitate communication-efficient SGD for federated learning, we ask the following questions.~\textit{Can we achieve lower per-iteration communication cost than QSGD and still guarantee convergence? What costs (in gradient accuracy) we need to pay at extremely low communication cost?}
%communication-efficient SGD for

To answer the above questions, we propose~\textit{hyper-sphere quantization} (\textit{HSQ}), a general framework of gradient compression that can be configured to achieve various trade-offs between communication efficiency and gradient accuracy. At the high compression ratio end, HSQ achieves a communication cost of $O(\log d)$ and the variance bound of gradient is blown up by $d$, which is favorable for federated learning. With a per-iteration communication cost of $O(\sqrt{d}\log d)$, HSQ achieves the same variance scaling as QSGD at $\sqrt{d}$. At the other extreme end with an $O(d)$ per-iteration communication cost, HSQ only increases the variance bound by a small constant. 

%The codewords in the codebook are vectors with the same dimension as $g$ and reside on the unit hyper-sphere.
Inspired by vector quantization techniques~\cite{rq,opq,pq,multiscale}, HSQ adopts a paradigm that is fundamentally different from existing gradient compression algorithms~\cite{signsgd,terngrad}. Instead of quantizing each element of the gradient vector $g$ individually or relying on sparsity, HSQ quantizes $g$ as a whole using a vector codebook shared between user devices and the central coordinator. HSQ chooses a codeword to approximate $g$ in a probabilistic manner and achieves low communication cost by only sending the index of the selected codeword. We prove that HSQ converges for both smooth convex and non-convex cost functions by analyzing its variance bound. We also demonstrate that some existing algorithms (e.g., SignSGD~\cite{signsgd}, TernGrad~\cite{terngrad}) can actually be regarded as special cases of HSQ under specific configurations. Experiments on state-of-the-art neural networks show that model training with HSQ converges smoothly. In terms of the total amount of communication to train the networks to convergence, HSQ significantly outperforms SGD and existing gradient compression algorithms.

\noindent\textbf{Contributions} Our contributions are three-folds. First, we provide a new paradigm that quantizes gradient using vector codebook. Vector quantization (VQ) has been extensively studied~\cite{rq,pq,l2hash} and this work may inspire the adoption of many effective VQ techniques for gradient compression. Second, by providing a continuum of trade-offs between communication efficiency and gradient accuracy, HSQ can be used in a diverse set of scenarios and helps understand the relation between variance and compression ratio in gradient quantization. Third, HSQ reduces the state-of-the-art per-iteration communication cost from $O(\sqrt{d}\log d)$ to $O(\log d)$, which benefits federated learning. 

\noindent\textbf{Notations} We use plain lower-case letters for scalars and vectors, e.g, $x$. Matrices are denoted by bold upper-case letters, i.e., $\vec{X}$. $\Vert x \Vert$ is the Euclidean norm of vector $x$ while $\Vert x \Vert_1$ is the $\ell_1$-norm of vector $x$. $|x|$ is the absolute value of scalar $x$. We denote a vector with all zeros by $\mathbf{0}$.

\section{Related Work}\label{sec:related}  

%\subsection{}  
\noindent\textbf{Gradient Compression for Data-Center-Based Training} It is widely known that gradient communication could easily become the bottleneck of data-center-based distributed machine learning when the model is large~\cite{qsgd,1bitSGD}. To reduce communication cost, the most intuitive idea is to transmit gradients with reduced precision.~\textit{TernGrad}~\cite{terngrad} quantizes each element of a gradient vector to three numerical values $\{-1, 0, 1 \}$. The gradient approximation $\tilde{g}$ is shown to be unbiased and training converges under the assumption of a bound on gradients. \textit{SignSGD}~\cite{signsgd} quantizes each element of gradient $g$ to $\{-1, 1 \}$ according to its sign. Although its gradient approximation is biased, SignSGD converges with respect to the $\ell_1$-norm of the gradients. Both TernGrad and SignSGD achieve an $O(d)$ communication cost at best as they quantize each dimension of a gradient vector individually.

\textit{QSGD}~\cite{qsgd} scales gradient vector $g$ by its Euclidean norm $\Vert g \Vert$ and quantizes each element in $g/\Vert g \Vert$ independently using $s$ uniformly spaced levels in $[0, 1]$. As there are at most $s(s+\sqrt{d})$ non-zero elements (in expectation) in the quantized gradient, QSGD achieves a per-iteration communication cost at $O(\sqrt{d}\log d)$ with $s=1$. \textit{GradiVeQ}~\cite{gradiveq} compresses the gradient $g$ using its projections on the eigenvectors (that corresponds to large eigenvalues) of the gradient covariance matrix. However, a training phase that communicates uncompressed gradients is required to learn the eigenvectors and GradiVeQ does not come with a theoretical analysis. Similar to QSGD and GradiVeQ, ATOMO~\cite{wang2018atomo} also relies on sparsity. It proposes the atomic decomposition on gradient $g$ and reduces communication cost by only sending atoms with non-zero (or large) weights. HSQ is different from these algorithms as it neither quantizes each element of the gradient independently nor relies on sparsity. It is difficult to apply the error feedback based algorithms~\cite{deepgrad,error_feedback} to federated learning as a user may not be selected in successive iterations and the error feedback may be stale. Moreover, HSQ is orthogonal to and can be combined with these algorithms.       

%\subsection\textbf{Gradient Compression for Federated Learning}

\noindent\textbf{Gradient Compression for Federated Learning} As communication cost is more critical for federated learning than data-center-based distributed training~\cite{fedAverage},  heuristics were proposed to achieve even lower per-iteration communication cost than the algorithms listed in Table~\ref{tab:cost}. Kone{\v{c}}n{\`y} et al.(~\citeyear{fedlearning}) proposed \textit{structured update} and \textit{random selection}. Structured update constrains the local gradient matrix to be low-rank, i.e., $\vec{W}=\vec{U}\vec{V}$, such that the low-rank matrices $\vec{U}$ and $\vec{V}$ can be reported with lower cost than $\vec{W}$. Random selection randomly chooses some elements in a gradient vector to report.~\textit{Selective SGD}~\cite{selectiveSGD} communicates elements in gradient $g$ with the largest absolute values. Although these heuristics may work in practice, they do not come with theoretical analysis and it is not clear how they trade gradient accuracy for communication efficiency. With an explicit variance bound, HSQ allows to adjust the balance between communication efficiency and gradient accuracy.

%HSQ achieves comparable or even lower communication cost than these heuristics~\footnote{This is because simply reporting the index of one selected gradient element in selective SGD costs $\log d$.}.
%\subsection{Vector Quantization}

\noindent\textbf{Vector Quantization} Vector quantization (VQ) techniques~\cite{l2hash}, including product quantization (PQ)~\cite{pq}, optimized product quantization (OPQ)~\cite{opq} and residual quantization (RQ)~\cite{rq}, are widely used to compress large datasets and conduct efficient similarity search. To compress a dataset containing vectors in $\mathbb{R}^d$, VQ uses one (or several) vector codebook(s) $\vec{C}=\begin{bmatrix} c_1 &c_2 &\cdots &c_{K}\end{bmatrix}\in \mathbb{R}^{d \times K}$, in which each codeword (column) $c(k)\in\mathbb{R}^d$. For a vector $x$, only the index of its nearest codeword in the codebook (i.e., $k^{\star}_x=\arg\min_{k} \Vert c(k)-x\Vert^2$) is stored. As the codebook $\vec{C}$ is shared over the entire dataset, the storage cost for each vector is reduced from $32d$ to $\log K$.

Although gradient is inherently a vector, directly applying VQ techniques to gradient compression is difficult. First, all VQ algorithms need to learn the codebook on the dataset to be compressed (e.g., using the kmeans algorithm). However, we can not get the gradients for codebook learning before model training starts. Second, the VQ algorithms do not provide guarantee on the quality of their approximations, which makes convergence analysis difficult. 

%In HSQ, we require the codewords to reside on the unit hyper-sphere and quantize the direction and norm of a gradient separately to avoid codebook training. Besides selecting the nearest codeword in the codebook greedily, an unbiased version of HSQ selects the codewords in a probabilistic manner to ensure that the gradient approximation is unbiased.           
                 
\section{Hyper-Sphere Quantization}\label{sec:hsq}

We consider the typical setting of federated learning with a central coordinator (or server) and many participating devices (or clients). Each device computes gradients using local training samples and transmits the (compressed) gradients to the coordinator. The coordinator aggregates the gradients from the devices and sends the model updates back to them. We assume that the devices use HSQ for gradient reporting while the coordinator transmits uncompressed model updates to devices in the analysis. In our experiments, model training also converges when the coordinator uses HSQ to transmit model updates.

\subsection{HSQ Algorithm}

HSQ partitions the original $d$-dimensional gradient vector $g$ into segments with length $d'$ (assume $d$ is divisible by $d'$) and quantizes each segment individually. For simplicity of presentation, we also use $g$ to denote a gradient segment. HSQ uses a vector codebook 
\[\vec{C}=\begin{bmatrix} c_1 &c_2 &\cdots &c_{m}\end{bmatrix}\in \mathbb{R}^{d' \times m}\] with $m$ codewords, in which each codeword (column) is a $d'$-dimensional unit-norm vector, i.e., $c_i \in \mathbb{R}^{d'}$ and $\Vert c_i \Vert \!=\! 1$ for $i \!=\! 1,\ldots, m$. We also require $\vec{C}$ to be a full-row-rank matrix, which means $m \ge d$ and normally $m = \Theta(d)$. ~\textit{Note that the codebook $\vec{C}$ is shared among the coordinator and the devices such that a device only needs to transmit the index of the selected codeword.} Moreover, the same codebook can be reused for all gradient segments. For a gradient segment $g \in \mathbb{R}^{d'}$, HSQ approximates it using a tuple $(\tilde{u}, c)$, in which $\tilde{u} \in \mathbb{R}$ is called the \textit{pseudo-norm} and $c$ is a codeword chosen from the codebook $\vec{C}$. Intuitively, $c$ encodes the unit-norm direction vector of $g$ while $\tilde{u}$ encodes the norm of $g$. Therefore, the HSQ-based approximation of $g$ is given by $\tilde{g}=\tilde{u}c$. We introduce two variants of HSQ, the unbiased version and the greedy version, in Algorithm~\ref{alg:unbaised-HSQ} and Algorithm~\ref{alg:greedy-HSQ}, respectively.

\begin{algorithm}[]
	\caption{Hyper-Sphere Quantization: Unbiased Version}
	\label{alg:unbaised-HSQ}
	\begin{algorithmic}
		\STATE {\bfseries Input:} a gradient segment $g \in \mathbb{R}^{d'}$ to compress
		\STATE {\bfseries Output:} tuple $(\tilde{u}, c)$ to approximate $g$
		\IF{$\Vert g \Vert = 0$}
		\STATE $u=0$, $c=c_1$, return
		\ENDIF
		\STATE Calculate $p=  \vec{C^\dagger} g$ where $\vec{C^\dagger}=\vec{C}^T(\vec{C}\vec{C}^T)^{-1}$
		\STATE Get $\tilde{p} \in \mathbb{R}^{d'}$ such that $\tilde{p}_i\!=\!\frac{|p_i|}{\Vert p \Vert_1}$, for $i\!=\!1,\ldots,d'$ 
		\STATE Select codeword $c_i$ as $c$ with probability $\tilde{p}_i$, if $c_i$ is selected, set $u=sign({p}_i)\cdot \Vert p \Vert_1$
		\STATE Quantize $u$ to obtain $\tilde{u}$ as $\tilde{u}=q(u)$ 
	\end{algorithmic}
\end{algorithm}

Algorithm~\ref{alg:unbaised-HSQ} gives the unbiased version of the HSQ algorithm, which chooses the tuple $(\tilde{u}, c)$ to approximate $g$ in a probabilistic manner. We will show that the gradient approximation $\tilde{g}=\tilde{u}c$ provided by Algorithm~\ref{alg:unbaised-HSQ} is unbiased in Section~\ref{sec:analysis-uHSQ}. Note that if a gradient segment is an all-zero vector, we set $u=0$ and $c=c_1$. A special configuration of Unbiased-HSQ is that $m=d$ and $\vec{C}$ is an orthonormal matrix, which means $\vec{C^\dagger}=\vec{C}^T$ and $p=\vec{C}^Tg$. In this case, codeword $c_i$ is selected as $c$ with a probability proportional to its correlation with $g$.   

There are $d/d'$ gradient segments in total and we assume that the minimum value and maximum value of $u$ in these segments are $u_{min}$ and $u_{max}$, respectively. Each $u$ is quantized to one of the $s+1$ uniformly spaced levels between $u_{min}$ and $u_{max}$ (inclusive). The quantization function $q(u)$ is similar to the one in QSGD and can be expressed as
\begin{eqnarray}\label{equ:norm}
q(u) \!=\!\!\!
&\!\begin{cases}
u_{min}\!+\!k\delta     & \text{with probability}~p(u) \\
u_{min}\!+\!(k\!+\!1)\delta & \text{otherwise}
\end{cases},
\end{eqnarray}  
where $p(u) = \frac{(k+1)\delta +u_{min}-u}{\delta}$, $\delta = (u_{max}-u_{min})/s$ and $u \in [u_{min}+k\delta,u_{min}+(k+1)\delta]$ with $k=0,\ldots, s-1$. Therefore, HSQ takes $\log(s+1) + \log(d')$ bits to communicate a gradient segment $g \in \mathbb{R}^{d'}$, in which $\log(s+1)$ bits are used for the pseudo-norm and $\log(d')$ bits are used to transmit the index of the selected codeword. Denote the HSQ quantized gradient of the $j$-th device as $(\tilde{u}^{j}, c^{j})$~\footnote{A device also sends  $u_{min}$ and $u_{max}$ to the coordinator so that $\tilde{u}^j$ can be decoded.}, the coordinator simply aggregates the gradients as $\bar{g} =\frac{1}{n} \sum_{j=1}^{n}\tilde{u}^jc^j$, where $n$ is the total number of participating user devices. We call SGD that uses HSQ for gradient reporting HSQ-SGD.

\begin{algorithm}[]
	\caption{Hyper-Sphere Quantization: Greedy Version}
	\label{alg:greedy-HSQ}
	\begin{algorithmic}
		\STATE {\bfseries Input:} a gradient segment $g \in \mathbb{R}^{d'}$ to compress
		\STATE {\bfseries Output:} tuple $(\tilde{u}, c)$ to approximate $g$
		\IF{$\Vert g \Vert = 0$}
		\STATE $u=0$, $c=c_1$, return
		\ENDIF
		\STATE Calculate the correlation vector $p=\vec{C}^T g$ 
		\STATE Select codeword $c_i$ as $c$, where $c_i = \arg\max_{c \in \vec{C}} |g^T c|$
		\STATE set $u = g^T c$ 
		\STATE Quantize $u$ to obtain $\tilde{u}$ as $\tilde{u}=q(u)$ using \eqref{equ:norm} 
	\end{algorithmic}
\end{algorithm}

Algorithm~\ref{alg:greedy-HSQ} gives the greedy version of the HSQ algorithm. The difference from Algorithm~\ref{alg:unbaised-HSQ} is that codeword selection is no longer probabilistic and the codeword that has the largest correlation with the gradient $g$ is chosen. Although the gradient approximation $\tilde{g}=\tilde{u}c$ provided by Greedy-HSQ is biased, we show that training converges with a growing epoch size in Section~\ref{sec:analysis-gHSQ}. Unbiased-HSQ and Greedy-HSQ have the same per-iteration communication cost but Greedy-HSQ usually performs better in the experiments.     

The paradigm of HSQ, which approximates the gradient vector with a direction vector and a pseudo-norm, is quite general. Several existing gradient compression algorithms, such as SignSGD and TernGrad, can actually be viewed as special cases of HSQ with a specific configuration of the segment length $d'$, codebook $\vec{C}$ and the method of pseudo-norm quantization. We provide more discussion about the relation between HSQ and these algorithms in Section 1 of the supplementary material.        

\subsection{Typical Configurations}\label{subsec:typical}

HSQ can achieve different trade-offs between communication efficiency and gradient accuracy by configuring the parameters, i.e., the length of gradient segment $d'$ and the number of quantization levels for pseudo-norm $s$. Based on the analytical results for Unbiased-HSQ in Section~\ref{sec:analysis-uHSQ}, we show three representative configurations~\textit{without pseudo-norm quantization}, which means that the exact $u$ is transmitted.  

\textbf{Extreme Compression} By quantizing the $d$-dimensional gradient vector as a whole (i.e., use a single segment with $d'=d$) and using an orthogonal matrix as $\vec{C}$ (i.e., $m=d$), HSQ uses $32 + \log(d)$ bits. This configuration achieves the current best per-iteration communication cost at $O(\log d)$ and blows up the variance bound of gradient by $d$. 
 
%$m=\Theta(d')$
\textbf{Compact Compression} With $d'=\sqrt{d}$ and $m=d'$, HSQ takes ($32\sqrt{d} + \frac{1}{2}\sqrt{d}\log d$) bits to transmit the entire gradient vector. The variance bound of gradient is scaled up by $\sqrt{d}$. This configuration resembles the sparse case of QSGD, which gives the previously known best communication cost of $O(\sqrt{d}\log d)$ with the same variance blow-up of $\sqrt{d}$. 

%$m=\Theta(d')$
\textbf{High Precision} Setting $d'=\kappa$, a small positive integer independent of $d$, and $m=d'$, HSQ has a communication cost of $O(d)$ and the variance bound of gradient is scaled up by $\kappa$. Under this configuration, HSQ has a communication cost similar to the algorithms that quantize each element in the gradient vector individually and the constant variance scaling resembles the dense configuration of QSGD.     

The three configurations show a clear trade-off between communication efficiency and gradient accuracy. A higher compression ratio leads to a larger variance bound on gradients, while reporting gradients more accurately incurs a higher communication cost. The number of user devices in federated learning is much larger than the number of machines in data-center-based distributed training. Variance can be reduced by averaging the gradients reported by a large number of devices rather than requiring each device to pay high per-iteration communication cost, which makes the extreme compression configuration of HSQ appealing for federated learning. 

\subsection{Codebook and Practical Considerations}

Consider the Greedy-HSQ in Algorithm~\ref{alg:greedy-HSQ}, the approximation error $\Vert g- uc\Vert^2=\Vert g\Vert^2-(g^T c)^2$ is small when $\beta(g, \vec{C})=\max_{c \in \vec{C}} |g^T c|$ is large. Note that the norm of $g$ only serves as scaling factor in the approximation error and we can assume $\Vert g\Vert\!=\!1$ without loss of generality. As we do not have any knowledge of $g$ when designing the codebook, we assume $g$ can appear anywhere on the unit hypersphere and optimize the worst case value of $\beta(g, \vec{C})$ , which leads to the following formulation of the codebook design problem    
\begin{equation}\label{codebook}
\vec{C}=\arg\max_{\vec{C}} \min_{y\in\mathbb{S}^{d'-1}} \max_{i = 1,\ldots, d'} |y^Tc_i|.
\end{equation} 
$\mathbb{S}^{d'-1}$ denotes the unit hypersphere in the $d'$-dimensional space. Although problem~\eqref{codebook} is difficult to solve, it requires that any vector on the unit hypersphere should have a codeword close to it in $\vec{C}$. Intuitively, the codewords in $\vec{C}$ should be uniformly located on the unit hypersphere such that the region covered by each codeword ($\mathbb{S}_i\!=\!\{y\in\mathbb{S}^{d'-1}:\Vert y-c_i\Vert\le\Vert y-c_j\Vert~\text{for}~j\!=\! 1,\ldots, m~\text{and}~j\neq i \}$) has identical area. When $m=d$, any orthonormal basis is uniformly located on the unit hypersphere and should be a good choice as codebook. Empirically, we observed that there is no difference in performance when setting $\vec{C}$ as different random rotations of the standard orthonormal basis. However, using the standard orthonormal basis provides slightly worse performance and this may be because the standard orthonormal basis only allows each device to update one element in a gradient segment. We discuss how to generate codewords uniformly located on the unit hypersphere when $m>d$ in Section 3 of the suppl. material. 

%% In Section~\ref{sec:anylysis}, we will show that HSQ can also use codebook containing $m$ ($m>d$) codewords and the codebook is not limited to orthonormal basis. 

Although the core idea of HSQ is a shared codebook between the coordinator and the devices, the coordinator does not need to really transmit $\vec{C}$ to the devices. Instead, a random seed can be issued to the devices to generate the codebook on their own to reduce communication. The overall complexity of 
matrix-vector multiplication ($\vec{C^\dagger} g$ or $\vec{C}^T g$) is $dm$ for all segments of a gradient vector and can be controlled by configuring $d'$ and $m$~\footnote{Remember that $m\ge d$ is required.}. When large $d'$ and $m$ are used for high compression ratio, complexity can be reduced in two ways. The first is to transform the matrices as $\bar{\vec{C}}=\frac{1}{\sqrt{k}}\vec{C^\dagger}\vec{H}$ or $\bar{\vec{C}}=\frac{1}{\sqrt{k}}\vec{C}^T\vec{H}$ and compute $\frac{1}{\sqrt{k}}\bar{\vec{C}}\vec{H}^Tg$ instead, in which $\vec{H}$ is an $d'\times k$ matrix ($k<d'$) and each entry follows i.i.d $\mathcal{N}(0,1)$. If the matrix transformations are conducted beforehand and $m=d$, the complexity to process a gradient vector is only $2dk$. According to the Johnson-Lindenstrauss lemma~\cite{vempala2005random}, this transformation preserves the inner product between $g$ and the rows of $\vec{C^\dagger}$ and $\vec{C}^T$ with high probability if $k$ is not too small. The other is using the standard orthonormal basis as the codebook, i.e., $\vec{C}=\begin{bmatrix} e_1 &e_2 &\cdots &e_{d'}\end{bmatrix}$, Unbiased-HSQ simply selects codeword $e_i$ with a probability proportional to $|g_i|$, while Greedy-HSQ selects the codeword $c_i$ with $i=\arg\max |g_i|$, both of which can be conducted very efficiently.

\section{Convergence Analysis}\label{sec:anylysis}

In this section, we present convergence results for both Alogrithm~\ref{alg:unbaised-HSQ} and Algorithm~\ref{alg:greedy-HSQ}. We begin our analysis by presenting the necessary definitions and assumptions. 
\begin{asmp_smooth}[$L$-smooth]
	\label{asmp_smooth}
	A function $f:\mathbb{R}^d \rightarrow \mathbb{R}$ is said to be $L$-smooth if for all $x, y \in \mathbb{R}^d$, it holds that
	\[
	f(x) \leq f(y) + \langle \nabla f(y), x-y \rangle + \frac{L}{2} \norm{x-y}^2.
	\]
\end{asmp_smooth}

\begin{asmp_cv}[Convex setting]
	\label{asmp_cv}
	The objective function $f(\cdot)$ is $L$-smooth and convex.
\end{asmp_cv}

\begin{asmp_ncv}[Non-convex setting]
	\label{asmp_ncv}
	The objective function $f(\cdot)$ is $L$-smooth but potentially non-convex.
\end{asmp_ncv}

\begin{asmp_vb}[Bounded second moment]
	\label{asmp_vb}
	For all $x\in \mathbb{R}^d$, let $g(x)$ denote the unbiased stochastic gradient at $x$, we require a $B'$-bounded second moment for all the segments $g'(x) \in \mathbb{R}^{d'}$ of $g(x)$, i.e., $\bE{\norm{g'(x)}^2} \leq B'$. Thus, the second moment of $g(x)$ is bounded by $\frac{d}{d'} B'$.
\end{asmp_vb}
\textit{Remark:} The assumption of bounded second moment is also required in QSGD (Definition 2.1). For a segment of stochastic gradient with second moment bound $B'$, the variance bound can be deduced as $\bE{\norm{g'(x)-\nabla f'(x)}^2}\le B'$ if $g'(x)$ is unbiased.   

%The proofs are provided in the supplementary material.  
 \subsection{Analysis for Unbiased-HSQ}\label{sec:analysis-uHSQ}
Now we present our main lemmas and theorems for Algorithm~\ref{alg:unbaised-HSQ}. The proofs can be found in Section 2 of the suppl. material. First, we show that the Unbiased-HSQ approximation $\tilde{g}=\tilde{u}c$ of a gradient segment $g$ is unbiased.
\begin{unbias}[Unbiasedness] \label{theorem-un-bias}
	Using the notations in Algorithm~\ref{alg:unbaised-HSQ}, given a full row rank codebook 
	$\vec{C} = \begin{bmatrix}
	c_1 &c_2 &\cdots &c_m
	\end{bmatrix} \in \mathbb{R}^{d' \times m}$
	, for any gradient segment $g\ne \vec{0} \in \mathbb{R}^{d'}$, we have that $\tilde{p}$ defines a probability over the indices $\{1, \ldots, m\}$  and the unbiasedness of $\tilde{g}$ with respect to the randomness of quantization, i.e., 
	$\E{\tilde{g}} = g.$
\end{unbias}
%\begin{proof}
%	Since $\M{C}$ has full row rank, we can denote its SVD as
%	$\M{C} = \M{U\Sigma V}^T = \M{U} \begin{bmatrix}\tilde{\M{\Sigma}} &0\end{bmatrix} \begin{bmatrix}\M{V}_1^T \\\M{V}_2^T \end{bmatrix}$, then 
%	$\M{C}^\dagger = \M{V}_1 \tilde{\M{\Sigma}}^{-1} \M{U}^T$, 
%	which implies that  $\M{C}^{\dagger T} \M{C}^\dagger = \M{U} \M{\Sigma}^{-2} \M{U}^T $ is positive definite. Recall that $p \triangleq \M{C}^\dagger g $, we have $\norm{p}^2 = g^T \M{C}^{\dagger T} \M{C}^\dagger g > 0$. Then, since for $i = 1,\ldots, m$, $\tilde{p}_i \triangleq \frac{|p_i|}{\norm{p}_1}$, we have that $\tilde{p}_i \in [0,1]$ and $\norm{\tilde{p}}_1 = 1$, which completes the proof of the first statement.
%	
%	For the unbiasedness of $\tilde{g} \triangleq \tilde{u} c$, first, it is clear that similar to QSGD (Lemma 3.1 (i)), we have the unbiasedness $\E{\tilde{u}} = u$ with respect to the randomness of $\tilde{u}$.
%	Then based on the construction of $\tilde{g}$, we have
%	\[ 
%	\begin{aligned}
%	\E{\tilde{g}} = \E{uc} &= \sum_{i=1}^{m} { \tilde{p}_i \cdot  sign(p_i) \cdot \norm{p}_1 \cdot c_i} \\
%	&= \M{C} p \\
%	&= \M{CC}^{\dagger} g \\
%	&= g,
%	\end{aligned}
%	\]which completes the proof.
%\end{proof}

Next, we analyze the variance brought by the quantization process of Unbiased-HSQ in Algorithm~\ref{alg:unbaised-HSQ}.  

\begin{comp_var}[Quantization Variance]\label{lemma:variance}
	Using the notations in Algorithm \ref{alg:unbaised-HSQ}, if Assumption \ref{asmp_vb} holds, the second moment of the quantized gradient segment $\tilde{g}\in \R^{d'}$ can be bounded as 
	\[
	\bE{\norm{\tilde{g}}^2} \leq m \sigma_1(\M{C}^\dagger)^2\cdot B' + \frac{(u_{max} - u_{min})^2}{s},
	\]where $\sigma_1(\M{C}^\dagger)$ denotes the largest singular value of $\M{C}^\dagger$. Moreover, with the quantized gradient at $x$ denoted as $\tilde{g}_x\in \R^d$, its variance can be upper bounded as
	\[
	\begin{gathered}
	\E{\norm{\tilde{g}_x - \nabla f(x)}^2} \leq \bE{\norm{\tilde{g}_x}^2}\\
	\leq V_q \triangleq \frac{d}{d'}\left(m \sigma_1(\M{C}^\dagger)^2\cdot B' + \frac{(u_{max} - u_{min})^2}{s}\right).
	\end{gathered}
	\]
\end{comp_var}

A variance bound that does not depend on $u_{max}$ and $u_{min}$ can also be obtained by relating the two values to the second moment of the gradient as
\[\bE{\norm{\tilde{g}}^2} \leq \left(1 + 4/s\right)m \sigma_1(\M{C}^\dagger)^2\cdot B'.\]
However, this bound is loose and we use the one related to $u_{max}$ and $u_{min}$ in the proof of convergence. Setting $m=d'$ and assuming that  $\vec{C}$ is an orthonormal matrix, Lemma~\ref{lemma:variance} can be re-rewritten as   
$
\bE{\norm{\tilde{g}}^2} \leq d'\cdot B' + (u_{max} - u_{min})^2/s.
$
%It also shows that we can reduce variance in the quantized gradient by dividing the gradient into shorter segments and paying more communication cost.

Several observations can be made from Lemma~\ref{lemma:variance}. First, the variances introduced by direction quantization and pseudo-norm quantization are additive. Second, if $\vec{C}$ is orthonormal and $m\!=\!d'$, the variance is blown up by $d'$ when the pseudo-norms are not quantized. This observation immediately leads to the configurations reported in Section~\ref{subsec:typical}. According to Lemma~\ref{lemma:variance}, one will not use more codewords than the length of a segment (i.e., $m \!>\! d'$) as this will blow up the variance bound. However, we observed empirically that increasing $m$ beyond $d'$ improves training performance. This is because the bound $ \norm{p}^2_1 \!\leq\! m \norm{p}^2$ used in the proof of Lemma~\ref{lemma:variance} is loose. With the variance bound in Lemma~\ref{lemma:variance}, we can obtain the following convergence results of Unbiased-HSQ.

%Lemma~\ref{lemma:variance} provides a principled way to allocate a fixed communication budget between direction and pseudo-norm quantization to minimize the variance in the quantized gradient.

\begin{SGD_C}[Convex, Theorem 6.3, \cite{bubeck2015convex}]
	\label{SGD_C}
	If Assumptions~\ref{asmp_cv}~and~\ref{asmp_vb} hold and using Lemmas~\ref{theorem-un-bias}~and~\ref{lemma:variance}, let $T$ be a positive integer and  $R^2=\norm{x_0-x^\star}^2$, where $x_0$ is the initial point of iterative scheme \textup{(1)}, choosing a constant step size $\eta_t = \eta = \frac{1}{L+1/\beta}$ with $\beta = \frac{R}{\sqrt{V_qT}}$, after running \textup{(1)} for $T$ iterations, we have the following inequality in expectation:
	\[ \E{ f\left(\frac{1}{T} \sum_{t=1}^{T}x_{t} \right) - f(x^\star) } \le R\sqrt{\frac{V_q}{T}} + \frac{L R^2}{2T}. \]
\end{SGD_C}

\begin{SGD_NC}[Non-convex]
	\label{SGD_NC}
	If Assumptions~\ref{asmp_ncv}~and~\ref{asmp_vb} hold and using Lemmas~\ref{theorem-un-bias}~and~\ref{lemma:variance}, let $T$ be a positive integer, $x_0$ be the initial point of iterative scheme \textup{(1)}, choosing a constant step size $\eta_t = \eta = \sqrt{\frac{2\left(f(x_0) - f(x^{\star})\right)}{TLV_q}}$, after running \textup{(1)} for $T$ iterations, we have the following inequality in expectation:
	\[ 
	\min_{0\le t \le {T-1}} \E{\norm{\nabla f(x_t)}^2} \leq \sqrt{\frac{2\left(f(x_0)-f(x^\star)\right)LV_q}{T}}.
	\]
\end{SGD_NC}

\subsection{Analysis for Greedy-HSQ}\label{sec:analysis-gHSQ}

 \begin{alpha_comp}[$\alpha$-Compressor]
 	\label{alpha_comp}
 	A compressor $Q(\cdot)$ is called an $\alpha$-Compressor if 
 	\[ (g^T Q(g))^2 \ge (1 - \alpha)||g||^2,\]
 	where $0 \le \alpha \le 1$ is a constant and $||Q(g)|| = 1$.
 \end{alpha_comp}
\begin{comp_err}[Quantization Error]\label{lemma:err}
	Compressor Q\[ Q(g) = \arg \max_{c_i \in \vec{C}} |g^Tc_i |  \]
	is an $\alpha$-compressor, where $g\in \mathbb{R}^{d'}$, $\vec{C}\in  \mathbb{R}^{d'\times m}$ and $0 < \alpha \le 1 - \sigma^2_{\min}(\vec{C}) / m$, $\sigma_{\min}(\vec{C})$ is the minimum singular value of the codebook matrix $\vec{C}$.
\end{comp_err}
Lemma~\ref{lemma:err} shows that the direction quantizer in Greedy-HSQ is an $\alpha$-compressor. Actually, we can substitute the direction quantizer in Greedy-HSQ with any $\alpha$-compressor and still preserve the convergence results in Theorem~\ref{theorem:greedy}. 

\begin{SGD_GD_NC}[Greedy-version, Non-convex]\label{theorem:greedy}
	\label{SGD_GD_NC} If Assumptions~\ref{asmp_ncv}~and~\ref{asmp_vb} hold, let the learning rate $\eta_t$ and batch size $n_t$ be 
	\[ 
	\eta_t = \frac{1}{\sqrt{T\mathcal{L}}} , \ \  
	n_t = \sqrt{T} ,\] 
	where $T > {\mathcal{L}} / {(1-\alpha)^2}$ is the total number of iterations, with $\mathcal{L} = L(1+4/s)d / d'$,
	$0<\alpha \le 1 - \sigma_{\min}(C) / m$, we have the convergence rate
	\[ 
	\sum_{t=0}^{T-1}\frac{1}{T}||g_t||^2 
	\le 
	\frac{(\alpha\sqrt{ T\mathcal{L}} + \mathcal{L}) {\sigma^2} / {\sqrt{T}} + 2 \mathcal{L}(f(x_0) - f(x_*))}{(1-\alpha) \sqrt{ T\mathcal{L}} - \mathcal{L}}. \]
\end{SGD_GD_NC}

The proof can be found in the supplementary material and the main technical challenge is that the gradient approximation given by an $\alpha$-compressor may be biased. The convergence rate of Greedy-HSQ is $O(1/\sqrt{T})$ but convergence is faster with small $\alpha$. This is intuitive as $\alpha$ models the error introduced by the compressor and smaller $\alpha$ means less error. Although convergence of Algorithm~\ref{alg:greedy-HSQ} relies on relatively large batch size (still smaller than SignSGD, where $n_t = T$) in Theorem~\ref{SGD_GD_NC}, empirically we observed that greedy-HSQ converges well with a small batch size.

%The proof is given in supplementary material, where the main contribution is that we show how to analyze convergence rate for any biased $\alpha$-compressor. The convergence rate of Greedy-HSQ is still $O(1/\sqrt{T})$ with one coefficient that is motonic to $\alpha$.  Since $\alpha$ represents the compressor quality, the smaller error compressor provides, the  smaller $\alpha$ is, the higher convergence rate we get. Although convergence of Algorithm~\ref{alg:greedy-HSQ} relies on relatively large batch size (still smaller than SignSGD, where $n_t = T$) in Theorem~\ref{SGD_GD_NC}, empirically we observed that greedy-HSQ converges well with a small batch size.
\section{Experimental Results}\label{sec:exp}

We experimented with popular deep neural networks, VGG~\cite{simonyan2014vgg19} and ResNet~\cite{he2016resnet} and report the performance of training image classifiers on ILSVRC-12~\cite{imagenet}, CIFAR-10, CIFAR-100~\cite{krizhevsky2009cifar10} and Fashion MNIST~\cite{fmnist} in the main paper and Section 6 of the suppl. material. The greedy version of HSQ is used due to its better empirical performance. The codebook $\vec{C}$ is generated by kmeans on random Gaussian vectors. Detailed experimental settings (e.g., learning rate scheduling and data augmentation) can be found in Section 4 of the suppl. material. We focus on the communication cost in federated learning and report it as the main performance metric. All codes will be made public.

\begin{figure}[!t]	
	\centering
	\includegraphics[width=0.49\columnwidth]{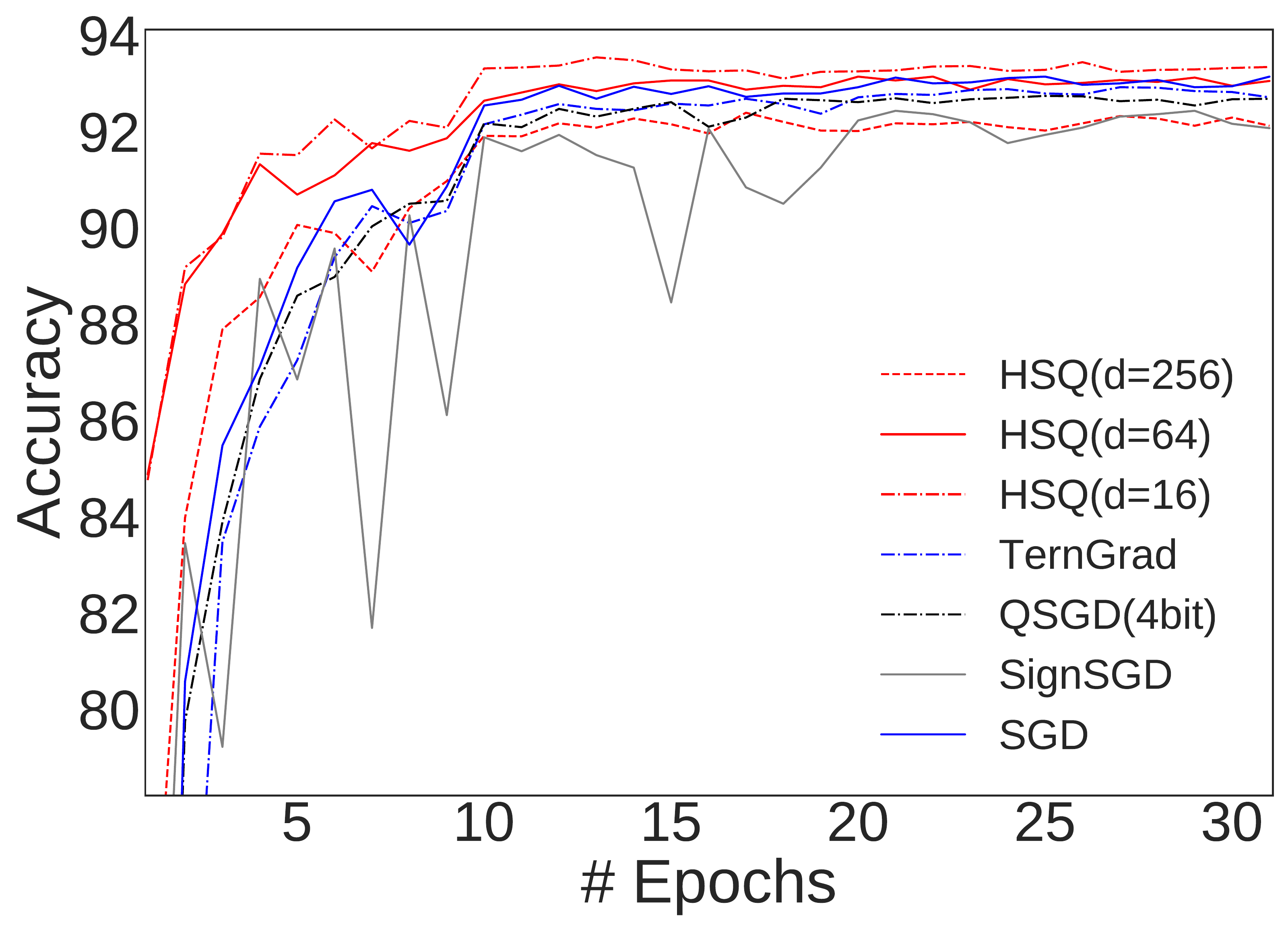}
	\includegraphics[width=0.49\columnwidth]{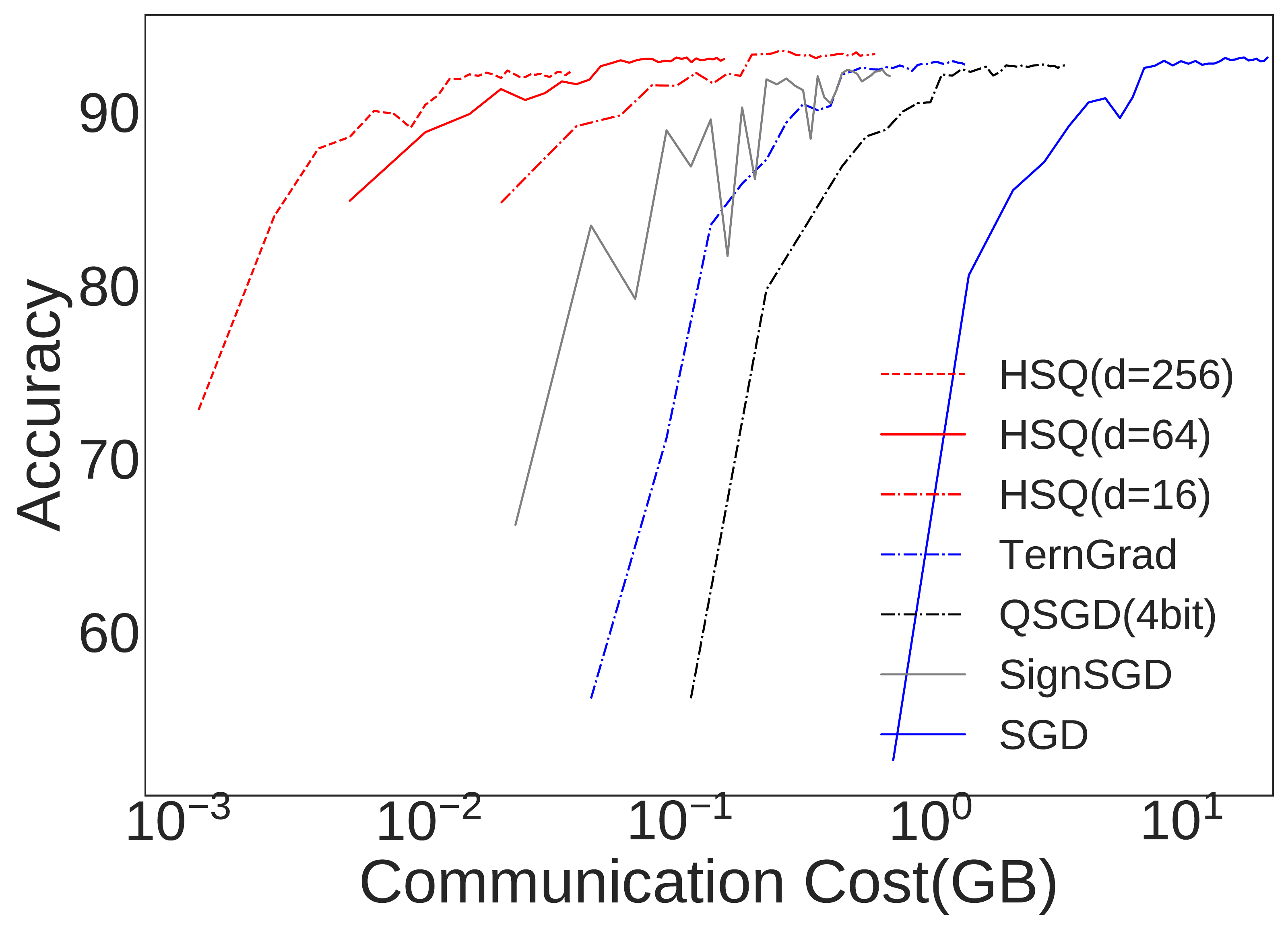} 
	\caption{Test accuracy vs. epoch (left) and communication cost (right) in federated learning setting for training ResNet50 on Fashion MNIST (best viewed in color)} 
	\label{fig:federated}
\end{figure}

\begin{figure}[!t]	
	\centering
	\includegraphics[width=0.49\columnwidth]{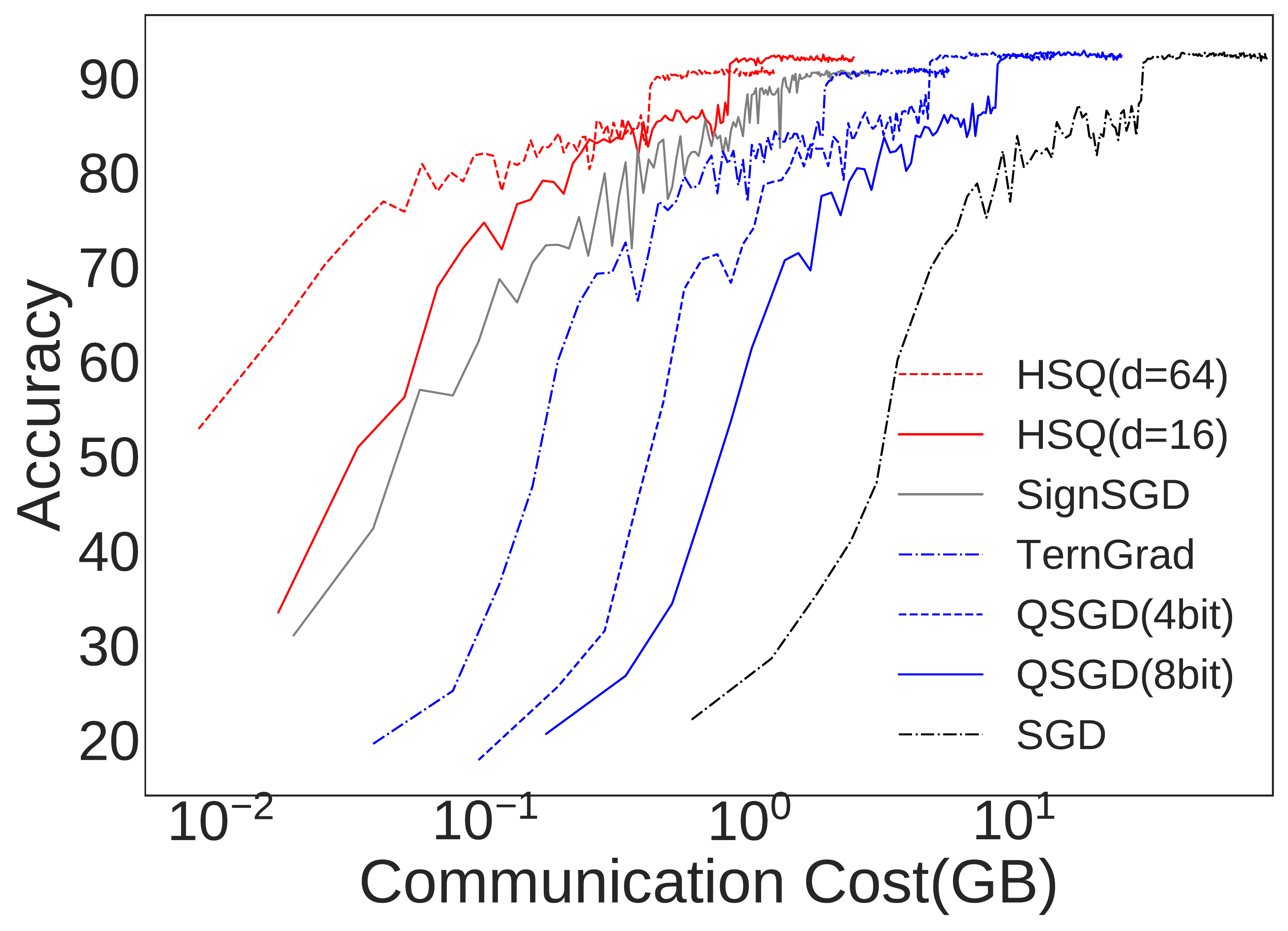}
	\includegraphics[width=0.49\columnwidth]{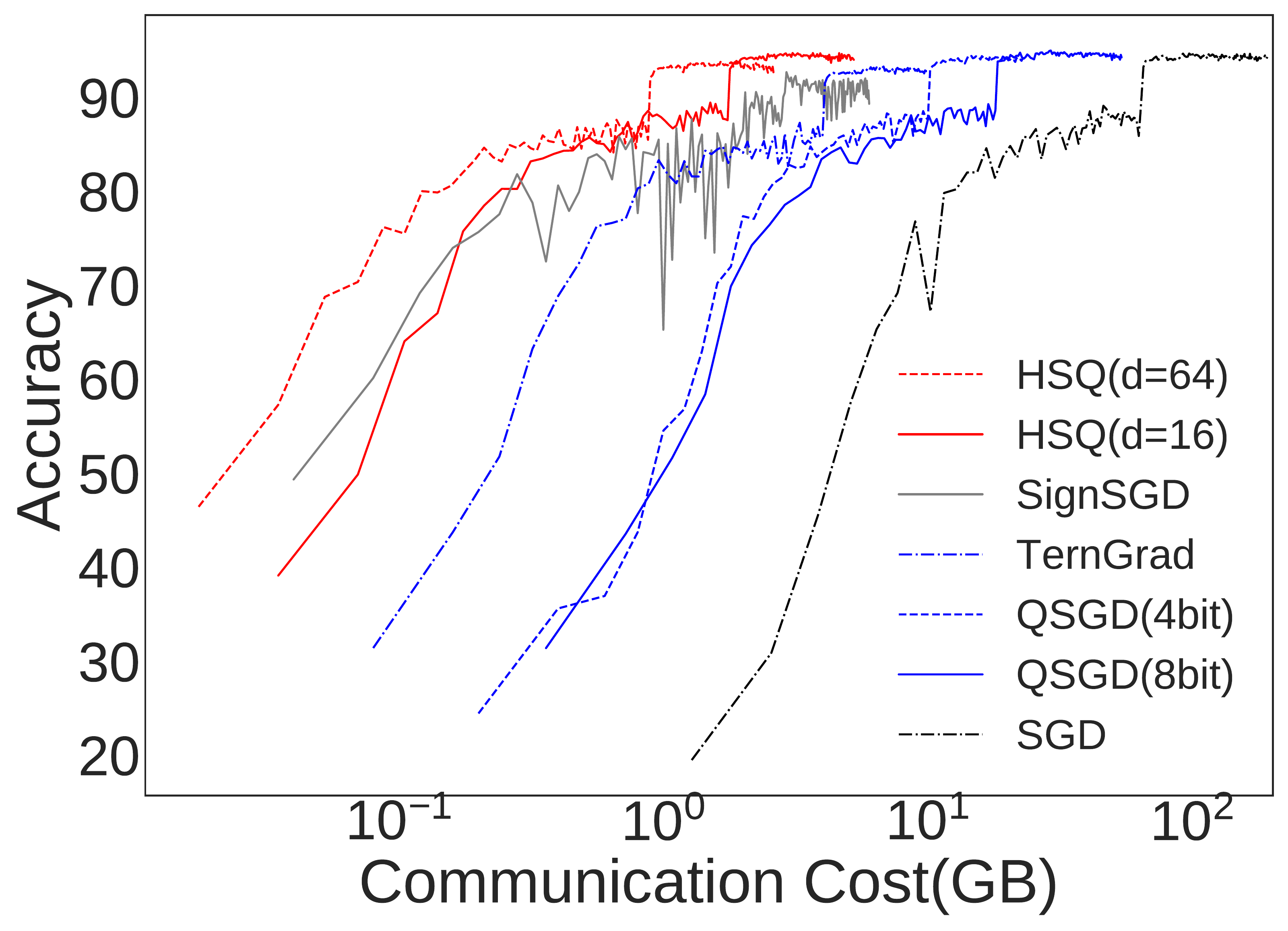}
	\caption{Test accuracy vs. communication cost for VGG19 (left)  and ResNet101 (right) training with SGD, QSGD, TernGrad, SignSGD and HSQ (best viewed in color)} 
	\label{fig:communication-accuracy}
\end{figure}

\begin{table*}[]
	\caption{Compression ratio and convergence accuracy of the algorithms on CIFAR10}
	\label{tab:result}
	\begin{center}
		%\begin{sc}
		\fontsize{7}{10}\selectfont
		\begin{tabular}{ccccccccc}
			\toprule
			Algorithm & SGD & SignSGD & TernGrad & QSGD (4 bit) & QSGD (8 bit)  & HSQ ($d$=8) & HSQ ($d$=16) & HSQ ($d$=64)\\ \midrule
			Comp. Ratio & 1 & 32 & 20.2 & $ \sim $8 & $ \sim $4 &  18.3 & 36.6 & 146.3  \\ \midrule
			VGG19&92.65 &90.79 &91.10 &92.60 &92.71 &\textbf{92.76} &92.38 &91.13 \\ \midrule
			ResNet50&94.19 &92.60 &93.29 &94.64 &94.03 &94.68 &\textbf{94.77} &93.77 \\ \midrule
			ResNet101&94.63 &92.01 &93.15 &94.35 &94.67 &94.48 &\textbf{94.70} &93.87 \\ \midrule
		\end{tabular}
		%\end{sc}
	\end{center}
\end{table*}

\noindent\textbf{Simulated experiments in federated learning setting} For experiments on Fashion-MNIST\cite{fmnist}, the training samples were randomly partitioned among 1000 users and 100 users are selected randomly for each iteration to simulate the scenario of federated learning. We report the test accuracy against epoch and the total amount of up-link communication conducted by user devices for gradient reporting in Figure~\ref{fig:federated}. We compared HSQ with SGD, SignSGD~\cite{signsgd}, TernGrad~\cite{terngrad}, and QSGD~\cite{qsgd}. The bucket size was set as 512 for QSGD as in~\cite{qsgd}. HSQ used 6 bits for the pseudo-norm and a codebook with 256 codewords for all configurations. The results show that training converges smoothly with HSQ and HSQ significantly reduces the amount of communication for achieving the same test accuracy comparing with the baselines. Setting $d\!=\!256$, HSQ reduces the communication cost of SGD by about 585x with only a loss of 0.8\% in final classification accuracy. With $d\!=\!64$ or $d\!=\!16$, HSQ achieves the same or slightly higher final classification accuracy compared with SGD but the communication cost is much lower.   

\begin{figure}[!t]	
	\centering
	\includegraphics[width=0.49\columnwidth]{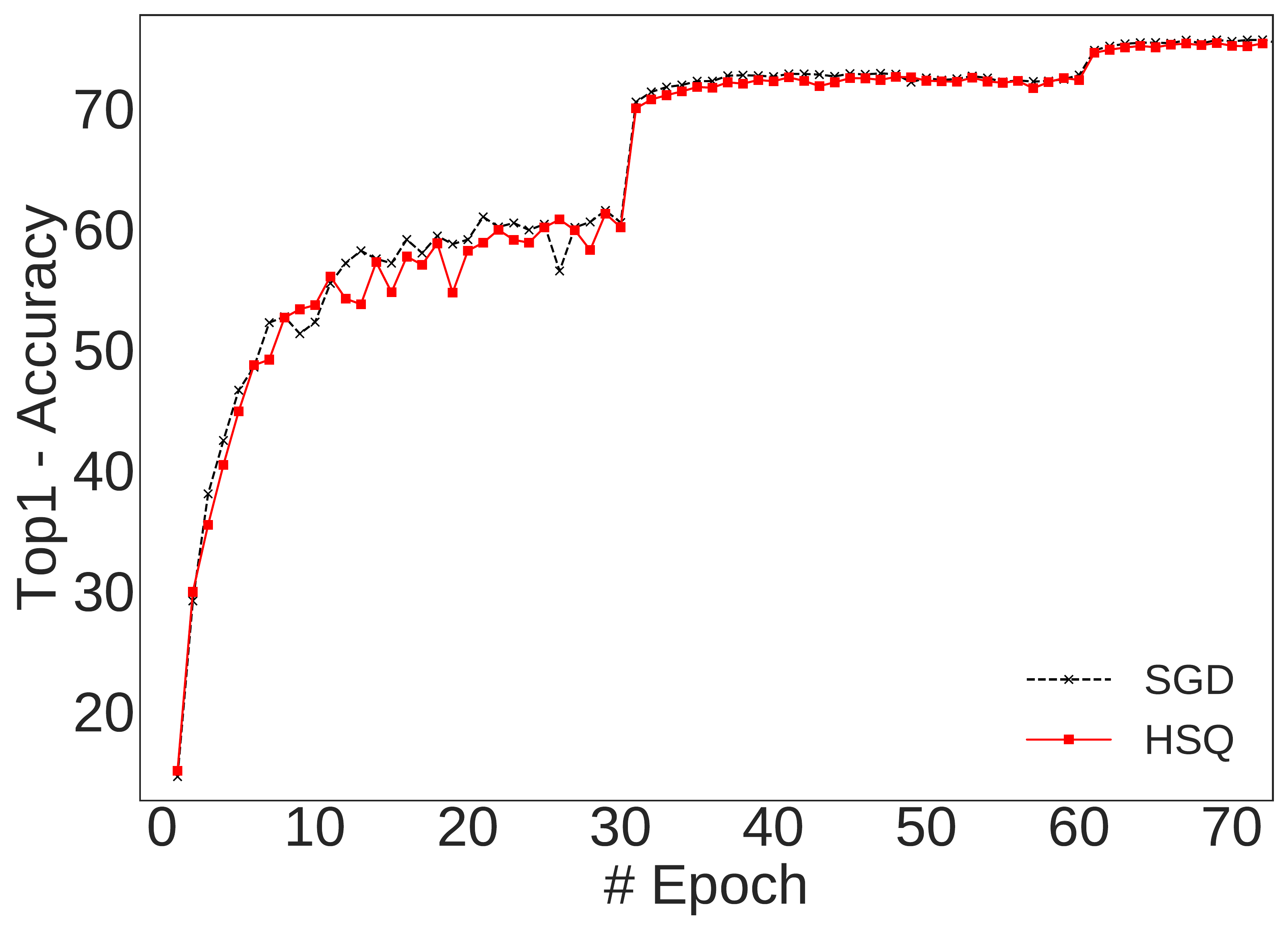}
	\includegraphics[width=0.49\columnwidth]{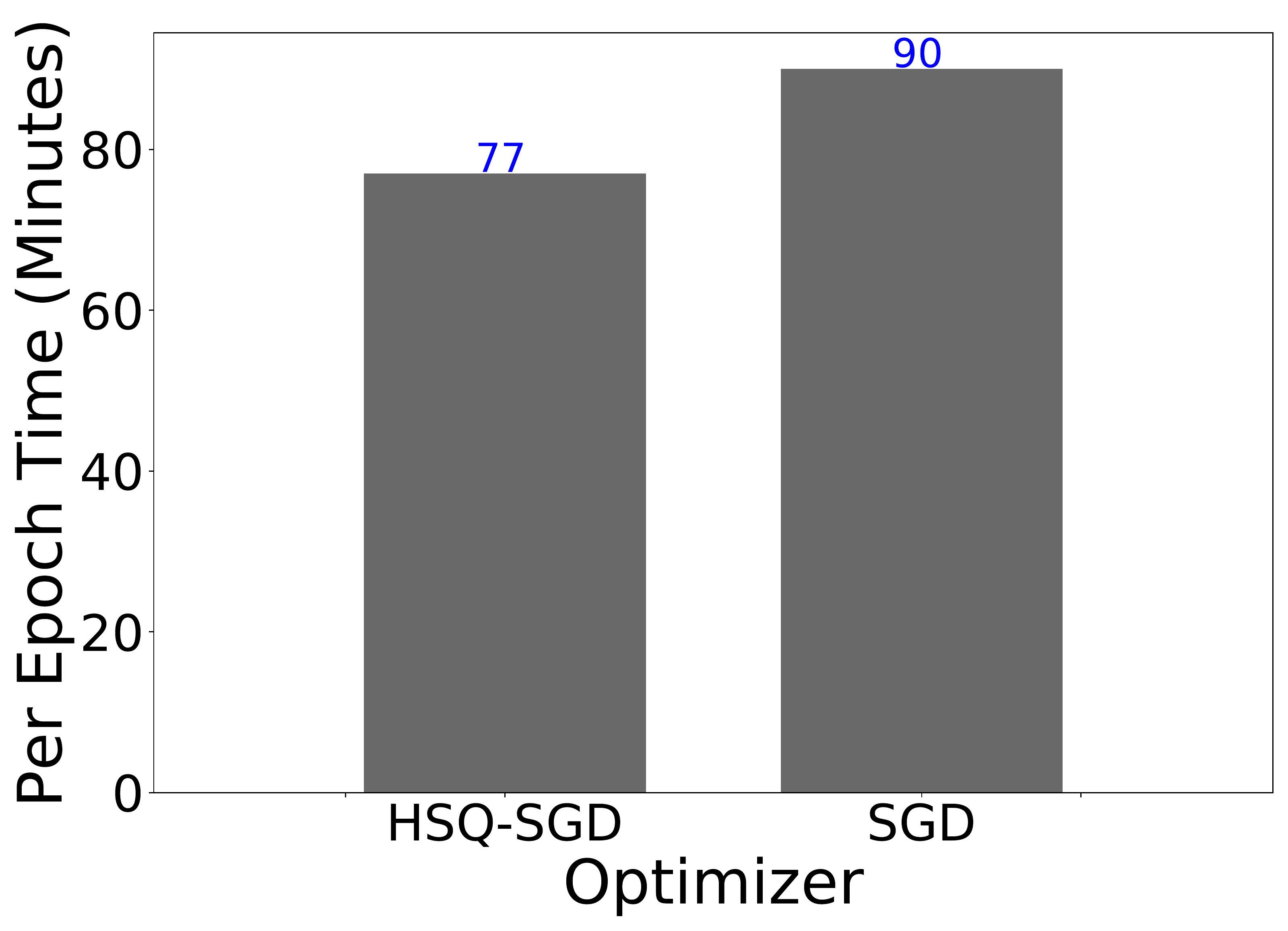} 
	\caption{Test accuracy vs. epoch (left) and per-epoch time (right) for SGD and HSQ (d=8) for training ResNet50 on ILSVRC-12 (best viewed in color)}
	\label{fig:distributed}
	\vspace{-3mm}
\end{figure}           

We conducted more experiments on CIFAR-10 and report the results in Figure~\ref{fig:communication-accuracy}. For clearer demonstration, we also list the compression ratio (compared with vanilla SGD as the baseline) and the convergence accuracy of the algorithms (with more configurations than shown in Figure~\ref{fig:communication-accuracy}) in Table~\ref{tab:result}. The results show that HSQ often outperform the baselines in both convergence accuracy and compress ratio. With $d\!=\!64$, the compression ratio of HSQ is significantly higher than the other algorithms  and the degradation in convergence accuracy is small. We plotted the test accuracy against the iteration count for the algorithms in Section 5.1 of the suppl. material, which shows training converges smoothly using HSQ.

\noindent\textbf{Timing experiments for distributed training} Although HSQ is designed for federated learning, where low communication cost is critical, we also report its performance for data-center-based distributed training in Figure~\ref{fig:distributed}. The dataset is ILSVRC-12~\cite{imagenet} and the 4 GPUs used for training are connected using high-speed PCIe bus. The results show that HSQ reduces the per-epoch time of SGD by 14.4\% due to smaller communication cost and the degradation in final test accuracy is very small ($<$0.5\%). 

\begin{figure}[!t]	
	\centering 
	\begin{subfigure}[b]{0.49\columnwidth}
		\includegraphics[width=\columnwidth]{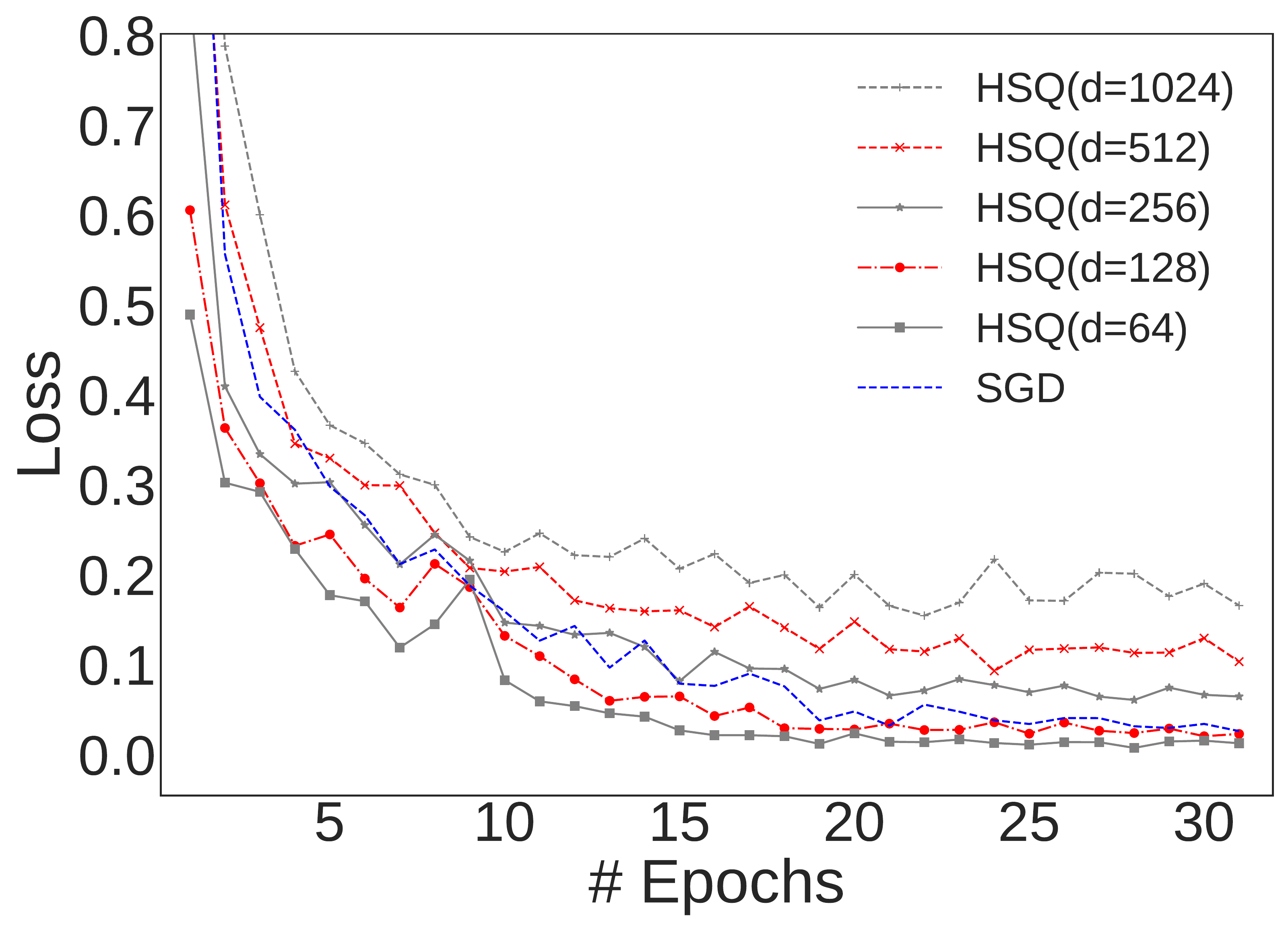}
		\caption{Value of $d$}
		\label{fig:loss-d}
	\end{subfigure}
	\begin{subfigure}[b]{0.49\columnwidth}
		\includegraphics[width=\columnwidth]{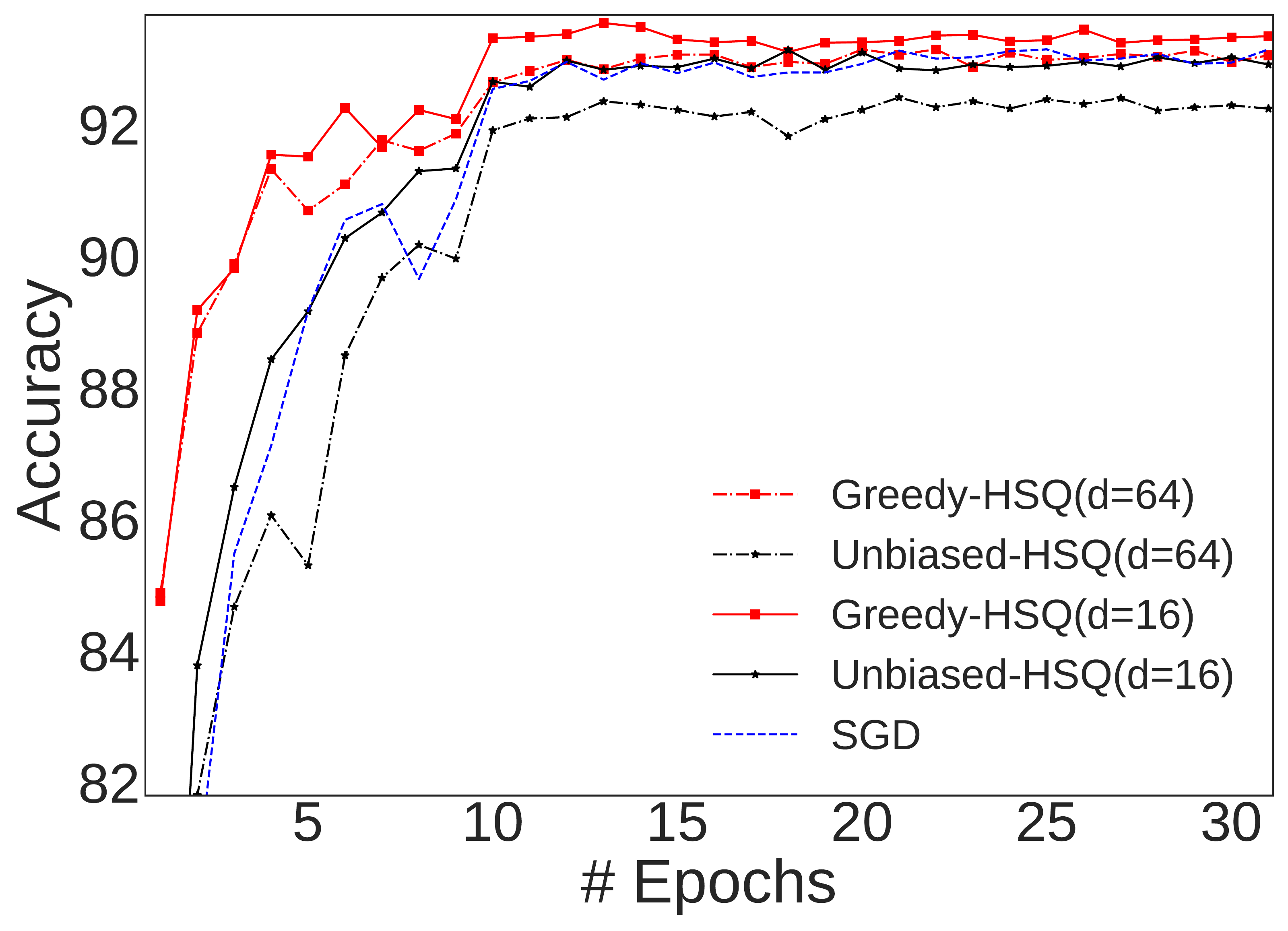}
		\caption{Greedy v.s. Unbiased HSQ}
		\label{fig:greedy vs. unbiased}
	\end{subfigure}
	\caption{Test of the parameter configurations in HSQ for training ResNet50 on Fashion-mnist (best viewed in color)} 
	\label{fig:para-fmnist} 	
\end{figure}

\begin{figure}[!t]	
	\centering 
	\begin{subfigure}[b]{0.49\columnwidth}
		\includegraphics[width=\columnwidth]{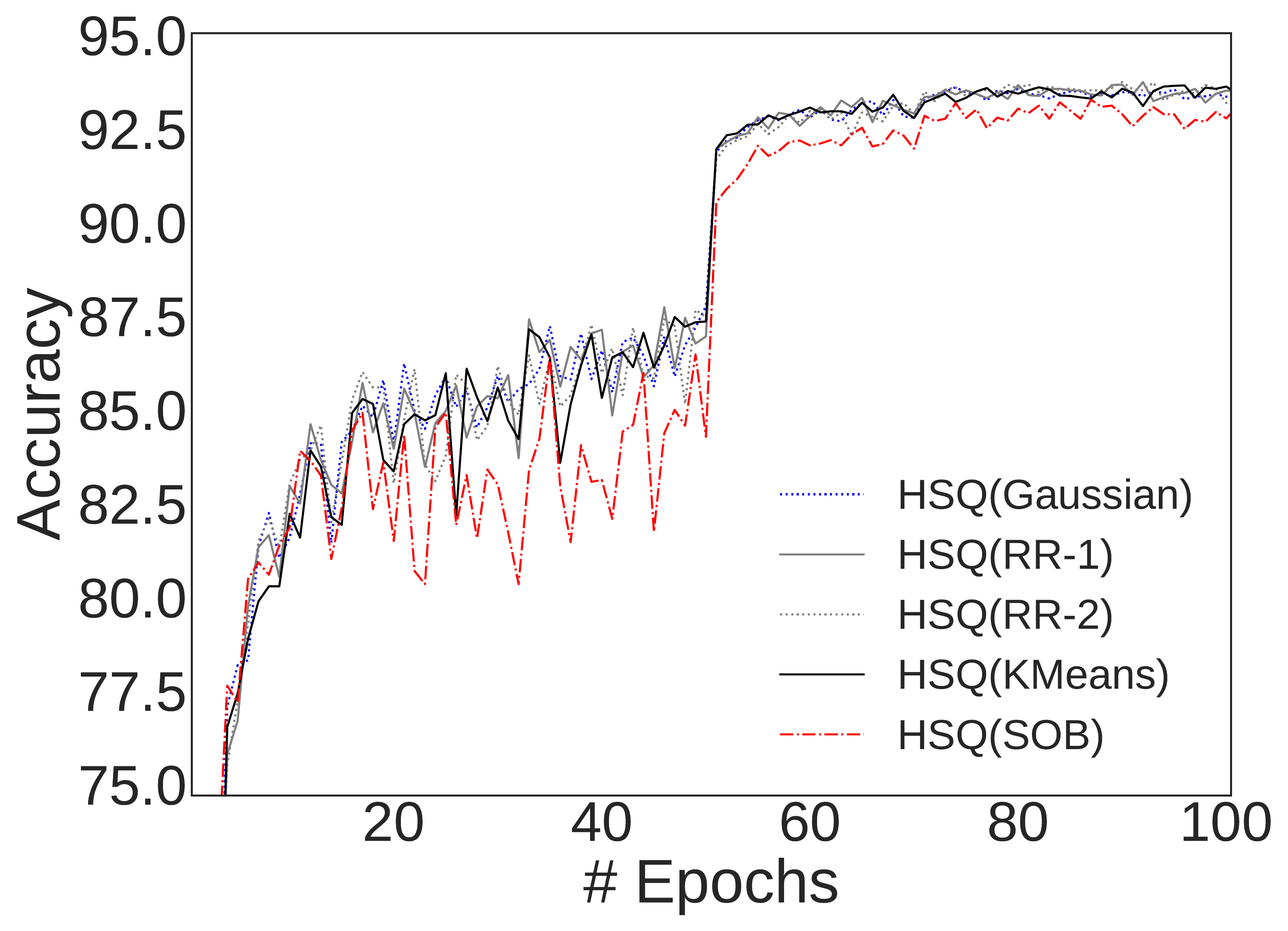} 
		\caption{Codebook generation}
		\label{fig:codebook}
	\end{subfigure}
	\begin{subfigure}[b]{0.49\columnwidth}
		\includegraphics[width=\columnwidth]{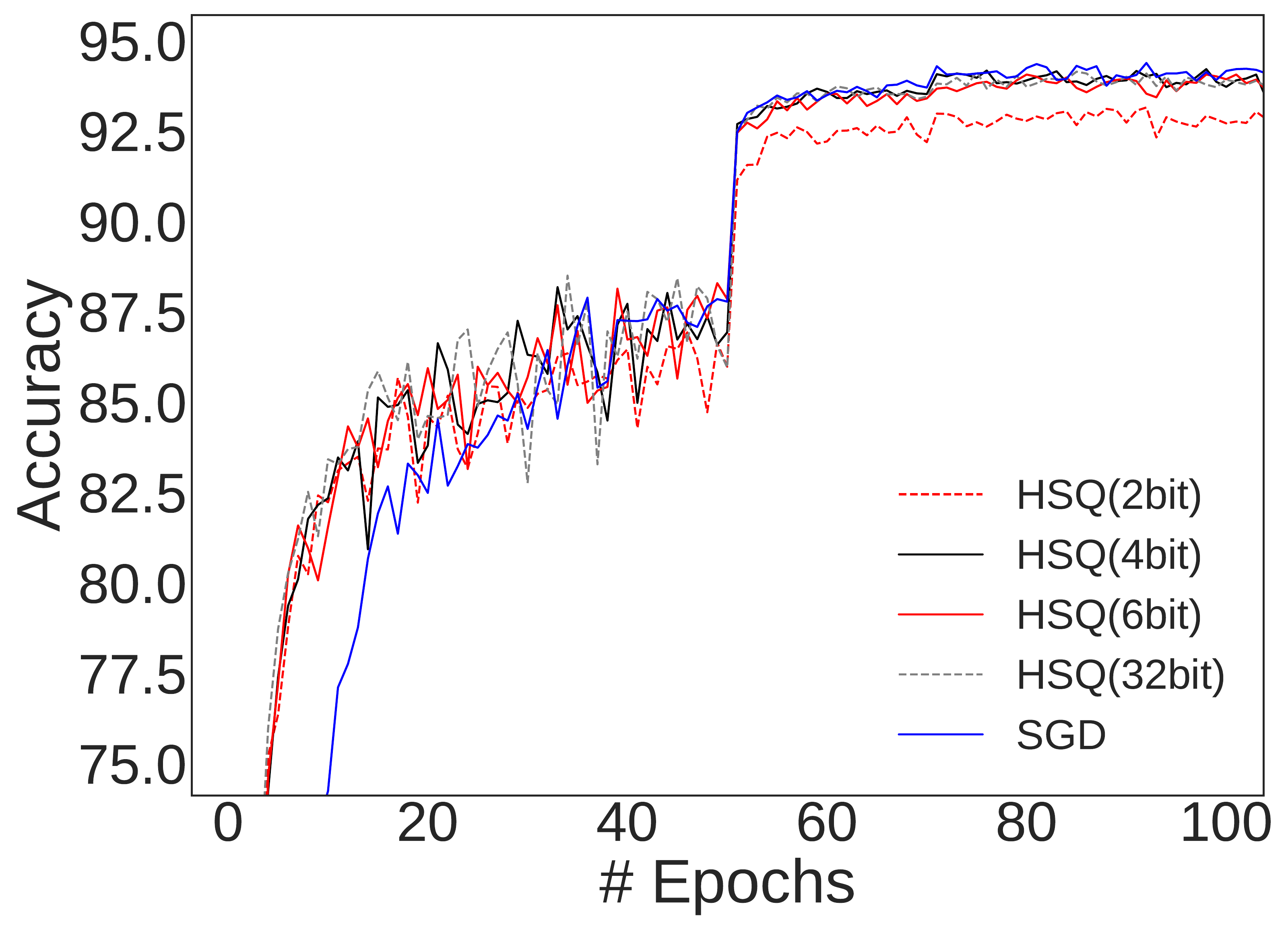}
		\caption{\#bits for $u$}
		\label{fig:bits for norm}
	\end{subfigure}
	\caption{Test of the parameter configurations in HSQ for training ResNet50 on CIFAR10 (best viewed in color)} 
	\label{fig:para-cifar} 	
	\vspace{-3mm}
\end{figure}

\noindent\textbf{Influence of the parameters} Keeping $m\!=\!d$, we plot the value of the loss function against epoch count in Figure~\ref{fig:loss-d} under different values of $d$. Note that larger $d$ results in higher compression ratio. The results show that when $d$ is too large (e.g., 512), the decrease of loss becomes unstable, which can be explained by the high variance in gradient. However, practical federated learning will involve a much larger number of users than we simulated in the experiments and averaging the gradients from different users reduces the variance. Thus, HSQ may use much larger $d$ (hence higher compression ratio) than we reported in the experiments in practical federated learning scenario. We compared Greedy-HSQ and Unbiased-HSQ in Figure~\ref{fig:greedy vs. unbiased}, which shows that Greedy-HSQ outperforms Unbiased-HSQ. Although the gradient approximation of Greedy-HSQ is biased, its performance is better possibly because the variance is smaller than Unbiased-HSQ.       

%We report the influence of the number of codewords on the performance in Figure~\ref{fig:loss-d}(b) with $d$ fixed at 32. The results show that increasing the number of codewords (using larger $m$ and paying more communication cost) provides better performance. 

We also tested different codebook generation methods, including standard orthonormal basis (SOB), the random rotation of SOB (RR), random Gaussian and K-means Gaussian, with $m=d=32$. Random Gaussian generates $m$ Gaussian vectors and normalizes them to unit norm. K-means Gaussian generates a large number of Gaussian vectors, conducts K-means with $m$ centers on them and normalizes the centers to unit norm. Figure~\ref{fig:codebook} shows that RR (RR-1 and RR-2 are two different random rotations), Gaussian and K-means Gaussian have almost the same performance while SOB performs slightly worse. Figure~\ref{fig:bits for norm} shows that using 4, 6 and 32 bits for pseudo-norm quantization provide almost the same performance but using only 2 bits hurts final test accuracy.           

\noindent\textbf{Additional experiments} Due to the page limit, we report additional experimental results in Section 5 of the suppl. material. For experiments in the main paper, the coordinator transmits uncompressed model updates. We show that the degradation in convergence accuracy is small (about 1\%) when the coordinator also uses HSQ to compress model updates. We also experimented the influence of the number of codewords (i.e., $m$) with $d$ fixed at 32. The results show that increasing the number of codewords (using larger $m$ and paying more communication cost) provides better performance.

\section{Conclusions}\label{sec:conclusion}  
We presented hyper-sphere quantization (HSQ), a general framework for gradient quantization that offers a range of trade-offs between communication efficiency and gradient accuracy via different configurations. HSQ achieves an extremely low per-iteration communication cost at $\log d$, where $d$ is the size of the model, and is guaranteed to converge for both smooth convex and smooth non-convex cost functions. The low per-iteration cost of HSQ is appealing for federated learning as it lowers the communication threshold to join the training and  encourages more users to participate. With HSQ, we demonstrate that vector quantization techniques can be effectively used for gradient compression. Given a rich literature of existing vector quantization techniques and that gradients are inherently high dimensional vectors, the idea of HSQ can stimulate more research along this direction.  

\bibliography{sphere}

\begin{thebibliography}{36}
\providecommand{\natexlab}[1]{#1}
\providecommand{\url}[1]{\texttt{#1}}
\expandafter\ifx\csname urlstyle\endcsname\relax
  \providecommand{\doi}[1]{doi: #1}\else
  \providecommand{\doi}{doi: \begingroup \urlstyle{rm}\Url}\fi

\bibitem[Acharya et~al.(2019)Acharya, De~Sa, Foster, and
  Sridharan]{sublinear_convex}
Acharya, J., De~Sa, C., Foster, D., and Sridharan, K.
\newblock Distributed learning with sublinear communication.
\newblock In \emph{International Conference on Machine Learning}, pp.\  40--50,
  2019.

\bibitem[Alistarh et~al.(2017)Alistarh, Grubic, Li, Tomioka, and
  Vojnovic]{qsgd}
Alistarh, D., Grubic, D., Li, J., Tomioka, R., and Vojnovic, M.
\newblock {QSGD: Communication-efficient SGD via gradient quantization and
  encoding}.
\newblock In \emph{Advances in Neural Information Processing Systems}, pp.\
  1709--1720, 2017.

\bibitem[Assran et~al.(2019)Assran, Loizou, Ballas, and Rabbat]{gradient_push}
Assran, M., Loizou, N., Ballas, N., and Rabbat, M.
\newblock Stochastic gradient push for distributed deep learning.
\newblock In \emph{International Conference on Machine Learning}, pp.\
  344--353, 2019.

\bibitem[Bernstein et~al.(2018)Bernstein, Wang, Azizzadenesheli, and
  Anandkumar]{signsgd}
Bernstein, J., Wang, Y.-X., Azizzadenesheli, K., and Anandkumar, A.
\newblock {signSGD: compressed optimisation for non-convex problems}.
\newblock \emph{arXiv preprint arXiv:1802.04434}, 2018.

\bibitem[Bubeck et~al.(2015)]{bubeck2015convex}
Bubeck, S. et~al.
\newblock Convex optimization: Algorithms and complexity.
\newblock \emph{Foundations and Trends{\textregistered} in Machine Learning},
  8\penalty0 (3-4):\penalty0 231--357, 2015.

\bibitem[Chen et~al.(2018)Chen, Choi, Brand, Agrawal, Zhang, and
  Gopalakrishnan]{adacomp}
Chen, C.-Y., Choi, J., Brand, D., Agrawal, A., Zhang, W., and Gopalakrishnan,
  K.
\newblock Adacomp: Adaptive residual gradient compression for data-parallel
  distributed training.
\newblock In \emph{Thirty-Second AAAI Conference on Artificial Intelligence},
  2018.

\bibitem[Chen et~al.(2010)Chen, Guan, and Wang]{rq}
Chen, Y., Guan, T., and Wang, C.
\newblock {Approximate nearest neighbor search by residual vector
  quantization}.
\newblock \emph{Sensors}, 10\penalty0 (12):\penalty0 11259--11273, 2010.

\bibitem[Ge et~al.(2013)Ge, He, Ke, and Sun]{opq}
Ge, T., He, K., Ke, Q., and Sun, J.
\newblock {Optimized product quantization for approximate nearest neighbor
  search}.
\newblock In \emph{Proceedings of the IEEE Conference on Computer Vision and
  Pattern Recognition}, pp.\  2946--2953, 2013.

\bibitem[He et~al.(2016)He, Zhang, Ren, and Sun]{he2016resnet}
He, K., Zhang, X., Ren, S., and Sun, J.
\newblock Deep residual learning for image recognition.
\newblock In \emph{Proceedings of the IEEE conference on computer vision and
  pattern recognition}, pp.\  770--778, 2016.

\bibitem[Jegou et~al.(2011)Jegou, Douze, and Schmid]{pq}
Jegou, H., Douze, M., and Schmid, C.
\newblock {Product quantization for nearest neighbor search}.
\newblock \emph{IEEE transactions on pattern analysis and machine
  intelligence}, 33\penalty0 (1):\penalty0 117--128, 2011.

\bibitem[Karimireddy et~al.(2019)Karimireddy, Rebjock, Stich, and
  Jaggi]{error_feedback}
Karimireddy, S.~P., Rebjock, Q., Stich, S., and Jaggi, M.
\newblock Error feedback fixes signsgd and other gradient compression schemes.
\newblock In \emph{International Conference on Machine Learning}, pp.\
  3252--3261, 2019.

\bibitem[Kone{\v{c}}n{\`y} et~al.(2016)Kone{\v{c}}n{\`y}, McMahan, Yu,
  Richt{\'a}rik, Suresh, and Bacon]{fedlearning}
Kone{\v{c}}n{\`y}, J., McMahan, H.~B., Yu, F.~X., Richt{\'a}rik, P., Suresh,
  A.~T., and Bacon, D.
\newblock {Federated learning: Strategies for improving communication
  efficiency}.
\newblock \emph{arXiv preprint arXiv:1610.05492}, 2016.

\bibitem[Krizhevsky \& Hinton(2009)Krizhevsky and
  Hinton]{krizhevsky2009cifar10}
Krizhevsky, A. and Hinton, G.
\newblock Learning multiple layers of features from tiny images.
\newblock Technical report, Citeseer, 2009.

\bibitem[Li et~al.(2014)Li, Andersen, Park, Smola, Ahmed, Josifovski, Long,
  Shekita, and Su]{paramtersever}
Li, M., Andersen, D.~G., Park, J.~W., Smola, A.~J., Ahmed, A., Josifovski, V.,
  Long, J., Shekita, E.~J., and Su, B.-Y.
\newblock {Scaling Distributed Machine Learning with the Parameter Server.}
\newblock In \emph{OSDI}, volume~14, pp.\  583--598, 2014.

\bibitem[Lin et~al.(2018)Lin, Han, Mao, Wang, and Dally]{deepgrad}
Lin, Y., Han, S., Mao, H., Wang, Y., and Dally, B.
\newblock Deep gradient compression: Reducing the communication bandwidth for
  distributed training.
\newblock In \emph{International Conference on Learning Representations}, 2018.
\newblock URL \url{https://openreview.net/forum?id=SkhQHMW0W}.

\bibitem[McMahan \& Ramage(2017)McMahan and Ramage]{federated}
McMahan, B. and Ramage, D.
\newblock {Federated learning: Collaborative machine learning without
  centralized training data}.
\newblock \emph{Google Research Blog}, 2017.

\bibitem[McMahan et~al.(2016)McMahan, Moore, Ramage, Hampson,
  et~al.]{fedAverage}
McMahan, H.~B., Moore, E., Ramage, D., Hampson, S., et~al.
\newblock {Communication-efficient learning of deep networks from decentralized
  data}.
\newblock \emph{arXiv preprint arXiv:1602.05629}, 2016.

\bibitem[Patarasuk \& Yuan(2009)Patarasuk and Yuan]{ring}
Patarasuk, P. and Yuan, X.
\newblock {Bandwidth optimal all-reduce algorithms for clusters of
  workstations}.
\newblock \emph{Journal of Parallel and Distributed Computing}, 69\penalty0
  (2):\penalty0 117--124, 2009.

\bibitem[Robbins \& Monro(1951)Robbins and Monro]{sgd}
Robbins, H. and Monro, S.
\newblock A stochastic approximation method.
\newblock \emph{Ann. Math. Statist.}, 22\penalty0 (3):\penalty0 400--407, 1951.

\bibitem[Russakovsky et~al.(2015)Russakovsky, Deng, Su, Krause, Satheesh, Ma,
  Huang, Karpathy, Khosla, Bernstein, Berg, and Fei-Fei]{imagenet}
Russakovsky, O., Deng, J., Su, H., Krause, J., Satheesh, S., Ma, S., Huang, Z.,
  Karpathy, A., Khosla, A., Bernstein, M., Berg, A.~C., and Fei-Fei, L.
\newblock {ImageNet Large Scale Visual Recognition Challenge}.
\newblock \emph{International Journal of Computer Vision (IJCV)}, 115\penalty0
  (3):\penalty0 211--252, 2015.
\newblock \doi{10.1007/s11263-015-0816-y}.

\bibitem[Sattler et~al.(2019)Sattler, Wiedemann, M{\"u}ller, and
  Samek]{non_iid_data}
Sattler, F., Wiedemann, S., M{\"u}ller, K.-R., and Samek, W.
\newblock Robust and communication-efficient federated learning from non-iid
  data.
\newblock \emph{arXiv preprint arXiv:1903.02891}, 2019.

\bibitem[Seide et~al.(2014)Seide, Fu, Droppo, Li, and Yu]{1bitSGD}
Seide, F., Fu, H., Droppo, J., Li, G., and Yu, D.
\newblock {1-bit stochastic gradient descent and its application to
  data-parallel distributed training of speech dnns}.
\newblock In \emph{Fifteenth Annual Conference of the International Speech
  Communication Association}, 2014.

\bibitem[Shokri \& Shmatikov(2015)Shokri and Shmatikov]{selectiveSGD}
Shokri, R. and Shmatikov, V.
\newblock {Privacy-preserving deep learning}.
\newblock In \emph{Proceedings of the 22nd ACM SIGSAC conference on computer
  and communications security}, pp.\  1310--1321. ACM, 2015.

\bibitem[Simonyan \& Zisserman(2014)Simonyan and Zisserman]{simonyan2014vgg19}
Simonyan, K. and Zisserman, A.
\newblock Very deep convolutional networks for large-scale image recognition.
\newblock \emph{arXiv preprint arXiv:1409.1556}, 2014.

\bibitem[Stich et~al.(2018)Stich, Cordonnier, and Jaggi]{sparsified}
Stich, S.~U., Cordonnier, J.-B., and Jaggi, M.
\newblock Sparsified sgd with memory.
\newblock In \emph{Advances in Neural Information Processing Systems}, pp.\
  4447--4458, 2018.

\bibitem[Suresh et~al.(2017)Suresh, Yu, Kumar, and McMahan]{distributed_mean}
Suresh, A.~T., Yu, F.~X., Kumar, S., and McMahan, H.~B.
\newblock Distributed mean estimation with limited communication.
\newblock In \emph{Proceedings of the 34th International Conference on Machine
  Learning-Volume 70}, pp.\  3329--3337. JMLR. org, 2017.

\bibitem[Tang et~al.(2018)Tang, Gan, Zhang, Zhang, and Liu]{decentralized}
Tang, H., Gan, S., Zhang, C., Zhang, T., and Liu, J.
\newblock Communication compression for decentralized training.
\newblock In \emph{Advances in Neural Information Processing Systems}, pp.\
  7652--7662, 2018.

\bibitem[Tang et~al.(2019)Tang, Yu, Lian, Zhang, and Liu]{double_squeeze}
Tang, H., Yu, C., Lian, X., Zhang, T., and Liu, J.
\newblock Doublesqueeze: Parallel stochastic gradient descent with double-pass
  error-compensated compression.
\newblock In \emph{International Conference on Machine Learning}, pp.\
  6155--6165, 2019.

\bibitem[Vempala(2005)]{vempala2005random}
Vempala, S.~S.
\newblock \emph{The random projection method}, volume~65.
\newblock 2005.

\bibitem[Wang et~al.(2018{\natexlab{a}})Wang, Sievert, Liu, Charles,
  Papailiopoulos, and Wright]{wang2018atomo}
Wang, H., Sievert, S., Liu, S., Charles, Z., Papailiopoulos, D., and Wright, S.
\newblock Atomo: Communication-efficient learning via atomic sparsification.
\newblock In \emph{Advances in Neural Information Processing Systems}, pp.\
  9850--9861, 2018{\natexlab{a}}.

\bibitem[Wang et~al.(2018{\natexlab{b}})Wang, Zhang, Sebe, Shen,
  et~al.]{l2hash}
Wang, J., Zhang, T., Sebe, N., Shen, H.~T., et~al.
\newblock {A survey on learning to hash}.
\newblock \emph{IEEE Transactions on Pattern Analysis and Machine
  Intelligence}, 40\penalty0 (4):\penalty0 769--790, 2018{\natexlab{b}}.

\bibitem[Wang et~al.(2019)Wang, Pi, and Zhou]{scalablebatch}
Wang, S., Pi, A., and Zhou, X.
\newblock Scalable distributed dl training: Batching communication and
  computation.
\newblock 2019.

\bibitem[Wen et~al.(2017)Wen, Xu, Yan, Wu, Wang, Chen, and Li]{terngrad}
Wen, W., Xu, C., Yan, F., Wu, C., Wang, Y., Chen, Y., and Li, H.
\newblock {Terngrad: Ternary gradients to reduce communication in distributed
  deep learning}.
\newblock In \emph{Advances in neural information processing systems}, pp.\
  1509--1519, 2017.

\bibitem[Wu et~al.(2017)Wu, Guo, Suresh, Kumar, Holtmann-Rice, Simcha, and
  Yu]{multiscale}
Wu, X., Guo, R., Suresh, A.~T., Kumar, S., Holtmann-Rice, D.~N., Simcha, D.,
  and Yu, F.
\newblock Multiscale quantization for fast similarity search.
\newblock In \emph{Advances in Neural Information Processing Systems}, pp.\
  5745--5755, 2017.

\bibitem[Xiao et~al.(2017)Xiao, Rasul, and Vollgraf]{fmnist}
Xiao, H., Rasul, K., and Vollgraf, R.
\newblock Fashion-mnist: a novel image dataset for benchmarking machine
  learning algorithms, 2017.

\bibitem[Yu et~al.(2018)Yu, Lin, Narra, Li, Li, Kim, Schwing, Annavaram, and
  Avestimehr]{gradiveq}
Yu, M., Lin, Z., Narra, K., Li, S., Li, Y., Kim, N.~S., Schwing, A., Annavaram,
  M., and Avestimehr, S.
\newblock {GradiVeQ: Vector Quantization for Bandwidth-Efficient Gradient
  Aggregation in Distributed CNN Training}.
\newblock In \emph{Advances in Neural Information Processing Systems}, pp.\
  5129--5139, 2018.

\end{thebibliography}
\bibliographystyle{icml2019}

\end{document}